\documentclass[conference,a4paper]{IEEEtran} 

\makeatletter
\def\thanks#1{\footnotemark\protected@xdef\@thanks{\@thanks
    \protect\footnotetext[\the\c@footnote]{#1}}}
\makeatother

\newcommand{\includegraphicsborder}[2][]{%
  \setlength{\fboxsep}{0pt}
  \fbox{%
    \includegraphics[#1]{#2}%
  }%
}

\usepackage{amsmath,amssymb,amsfonts}
\usepackage{algorithmic}
\usepackage{graphicx}
\usepackage{textcomp}
\usepackage{xcolor}

\usepackage[numbers,sort,compress]{natbib} 
\usepackage{cite}

\begin{document}

\title{Curriculum as Code: An AI-Assisted Architecture for Instructional Design in STEM Education}

\author{
{Henrique Mohallem Paiva, \textit{Senior Member, IEEE}}
\thanks{H. M. Paiva is with the Universidade Federal de Sao Paulo ({Unifesp}), Rua Talim, 330, Sao Jose dos Campos (SP), Brazil, 12231-280, and with the Institute of Technology and Leadership ({Inteli}), Av. Prof. Almeida Prado, 520, Sao Paulo (SP), Brazil, 05508-901. E-mail: hmpaiva@unifesp.br} 
\thanks{\newline \textit{This work has been submitted to the IEEE for possible publication. Copyright may be transferred without notice, after which this version may no longer be accessible.}}
}

\maketitle

\begin{abstract}
\textit{Contribution:} This paper presents a six-phase Artificial Intelligence (AI)-assisted instructional design architecture based on the Curriculum as Code paradigm. It integrates Generative AI with LaTeX and Python to automate the creation of reproducible, visually consistent, and technically precise materials for Science, Technology, Engineering, and Mathematics (STEM) education.
\textit{Background:} The creation of customized instructional materials, particularly for active learning environments, imposes a heavy workload on faculty. Standard presentation tools lack robust support for complex technical content, while current AI applications often suffer from hallucinations and fail to formalize the instructional authoring process, limiting their utility for rigorous academic design.
\textit{Intended Outcomes:} The proposed framework aims to reduce class preparation time while ensuring mathematical accuracy, strict adherence to institutional visual identity, and the preservation of the instructor's tacit pedagogical knowledge through explicit rules.
\textit{Application Design:} The solution consists of a structured six-phase pipeline that replaces ad-hoc prompt engineering with a systematic workflow. It utilizes text-based interfaces and code-driven generation (LaTeX/Beamer for structured slides and Python for technical figures), governed by explicit pedagogical constraints, contextual calibrations, and automated review cycles.
\textit{Findings:} Validated over a one-year implementation period across 8 distinct academic modules and 28 project contexts in a demanding Project-Based Learning (PBL) environment, the architecture was associated with a substantial decrease in instructor workload. The generated assets underwent independent peer review and were successfully deployed in the classroom by six different faculty members, confirming that the methodology captures tacit knowledge and scales beyond a single author. Furthermore, based on over 600 voluntary student evaluations, the materials achieved consistently high quality ratings, ranging from 8.5 to 9.9 out of 10. The results indicate high technical reproducibility, minimized AI hallucinations, and sustained pedagogical and visual fidelity, suggesting its viability for broad STEM educational applications.
\end{abstract}

\begin{IEEEkeywords}
Generative AI, Instructional Design, Active Learning, Engineering Education
\end{IEEEkeywords}

\renewcommand{\arraystretch}{1.3}

\section{Introduction}

Active learning methodologies, particularly Project-Based Learning (PBL) \citep{barros2023using,paiva2025pbl,hayashi2025simulation} and Flipped Classroom \citep{gong2024exploring,ye2026project}, have fundamentally redefined engineering education by shifting the instructor's role from a traditional transmitter of knowledge to a facilitator of learning. This pedagogical paradigm requires instructional materials to be highly customized, deeply anchored in practical project contexts, and deliberately designed to promote student autonomy and critical thinking.

Traditional visual authoring tools, commonly referred to as WYSIWYG tools (\textit{What You See Is What You Get}), provide intuitive interfaces for general-purpose documents and presentation materials but offer limited native support for complex mathematical notation, algorithmic pseudocode, and structured scientific content. As a result, instructors often spend considerable effort manually adjusting formatting and maintaining consistency across slides, reducing the time available for instructional design \citep{knauff2014efficiency}.

While the advent of Generative Artificial Intelligence (GenAI) offers promising avenues for educational technology, the current literature predominantly emphasizes isolated content generation \citep{das2025educator}, intelligent tutoring systems \citep{almetnawy2025adaptive}, and personalized learning paths \citep{tu2025empowering}. There is a noticeable gap regarding the formalization of instructional authoring processes. Specifically, there is little attention given to AI-assisted workflows that can systematically enforce pedagogical coherence, maintain institutional visual identity, and guarantee mathematical and technical reproducibility across an entire curriculum.

To address this gap, this paper investigates the following research questions (RQs):
\begin{itemize}
    \item \textbf{RQ1:} How can an AI-assisted instructional design workflow be structured to effectively translate an instructor's tacit pedagogical knowledge into explicit, replicable rules while preserving pedagogical consistency?
    \item \textbf{RQ2:} How does the integration of code-based authoring tools with Generative AI mitigate technical hallucinations and ensure the mathematical and visual reproducibility of STEM instructional materials?
    \item \textbf{RQ3:} What is the practical impact of this automated architecture on faculty preparation time and the maintenance of institutional visual identity when deployed in a highly contextualized learning environment?
\end{itemize}

It is important to note that the PBL model inherently adds a layer of complexity to instructional design due to the strict requirement of contextualizing theoretical concepts within practical projects. Because the proposed architecture successfully resolves this highly complex scenario, its application to traditional, lecture-based classes becomes more straightforward. 

The remainder of this paper is organized as follows. Sections II and III review the theoretical foundations and related work regarding active learning and Curriculum as Code. Section IV details the proposed AI-assisted instructional design workflow, including its phases and technical implementation using LaTeX and Python. Section V presents the validation and results gathered over a one-year implementation period. Section VI discusses the practical implications and future work, followed by the concluding remarks in Section VII.

\section{Related Work}

The application of Generative AI to education has been extensively explored in recent literature, with studies spanning intelligent tutoring systems \citep{maurya2025pedagogy}, automated assessment \citep{yavariabdi2025generative}, and personalized learning pathways \citep{bucchiarone2024scalable, drobnjak2026systematic}. However, the majority of these contributions focus on the student-facing side of education, supporting learners through dialogue, feedback, or adaptive content delivery. The instructor-facing side, particularly the automation of instructional material production, remains comparatively underexplored. This section reviews the existing body of work relevant to our proposal, organized around three axes: (i) the use of LLMs for generating educational content, (ii) structured prompt engineering frameworks for instructional design, and (iii) technical approaches that integrate code-based authoring tools with generative models.

Several recent studies have investigated the capacity of large language models to generate educational materials, such as lecture summaries \citep{hashiyada2025framework}, multiple-choice questions \citep{awalurahman2025transformer}, and explanatory texts for STEM topics \citep{kasneci2023chatgpt}. These works demonstrate that LLMs can produce content that is often coherent and contextually appropriate. However, they also consistently report significant limitations: hallucinations, factual inaccuracies, and a tendency to generate overly generic or verbose outputs that require extensive human revision \citep{adejumo2026systematic, chen2025mitigating}. In the context of quantitative disciplines, where mathematical notation and algorithmic precision are paramount, these limitations are particularly problematic. The lack of fine-grained control over the format, layout, and visual hierarchy of generated content further restricts the applicability of these approaches to professional instructional design.

To address these limitations, recent research has turned to prompt engineering as a mechanism for improving the reliability and specificity of LLM outputs. Studies have proposed techniques such as few-shot prompting, chain-of-thought reasoning, and role-playing to enhance performance in educational tasks \citep{qian2025prompt}. Nevertheless, most of these efforts remain \textit{ad hoc} and problem-specific, lacking a systematic architecture that can be replicated across different courses or pedagogical contexts. Notably, the concept of a multi-phase pipeline, where the model is first calibrated, then plans, and finally executes, has been proposed in general-purpose LLM applications \citep{lyu2025multiphase, matei2026automated, nag2026multi}, but its adaptation to instructional design, with explicit pedagogical constraints and visual consistency requirements, is absent from the current literature. Our work addresses this gap by proposing an end-to-end architecture that transforms prompt engineering from an ephemeral practice into a repeatable, curriculum-wide process.

A parallel body of work has explored the integration of code-based authoring tools with generative AI. The use of LaTeX and Beamer for academic slide production is well established \citep{silva2011model}, and recent efforts have investigated the feasibility of using LLMs to generate LaTeX code \citep{das2025educator, lyn2025translatex, kale2025texpert}. These studies report that, while LLMs can produce syntactically correct LaTeX for simple documents, they often fail when faced with custom document classes, complex macros, or specific package interactions \citep{lyn2025translatex, kale2025texpert}. Similarly, the generation of scientific figures through Python libraries such as Matplotlib and Seaborn has shown promising results for simple visualizations; however, the generated code often requires manual refinement to correctly configure visualization parameters and produce publication-quality figures \citep{khan2025evaluating}. Our architecture addresses this by separating the generation of Python code for figures from the LaTeX content, and by using few-shot calibration with instructor-provided examples to ensure that both outputs adhere to institutional templates and pedagogical rules.

To the best of our knowledge, no previous work has proposed a comprehensive, replicable architecture that simultaneously addresses pedagogical calibration, visual consistency, and technical reproducibility in the production of instructional materials for active learning in STEM. Existing frameworks either focus on content generation without enforcing structural or visual constraints, or they rely on manual post-processing that undermines the efficiency gains promised by automation. Our contribution is the formalization of a six-phase pipeline: Context Injection, Pedagogical Calibration, Technical Calibration, Planning, Iterative Implementation, and Review. It systematically embeds the instructor's tacit knowledge, applies Curriculum as Code principles, and enforces cognitive load theory constraints, resulting in a workflow that significantly reduces preparation time while preserving pedagogical and mathematical fidelity.

\section{Theoretical Foundation}

The proposed architecture is grounded on five interrelated theoretical and technical pillars: active learning methodologies, the current state and limitations of generative AI in education, the concept of Curriculum as Code, knowledge elicitation techniques, and cognitive load theory. Together, these foundations justify both the pedagogical rules embedded in the prompts and the technical choices regarding LaTeX and Python.

Active learning methodologies, particularly Project-Based Learning (PBL) and the Flipped Classroom, have reshaped engineering education by shifting the instructor's role from a mere transmitter of knowledge to a facilitator of autonomous learning \citep{barros2023using,paiva2025pbl,hayashi2025simulation, gong2024exploring,ye2026project}. In PBL, theoretical concepts must be consistently anchored in authentic, practical project contexts, requiring instructional materials that are not only mathematically precise but also deeply contextualized. Similarly, the Flipped Classroom model demands concise, reference-ready materials that students can study autonomously before class, freeing in-person time for problem-solving and discussion. These pedagogical approaches impose strict requirements on instructional design: the theory must be fragmented into digestible modules, directly connected to the project context, and immediately followed by applied activities. This structured flow (theory, key takeaways, activity, and answer) is precisely what our architecture seeks to enforce automatically.

The emergence of Generative Artificial Intelligence (GenAI), particularly large language models, has opened new possibilities for educational content creation. Current studies have explored their use in intelligent tutoring systems, personalized learning paths, and automated question generation. However, the literature still presents significant methodological limitations when GenAI is applied to the production of instructional materials for quantitative disciplines. Two critical issues are hallucinations (the generation of mathematically or factually incorrect content) and low reproducibility, where repeated requests for similar materials yield inconsistent outputs \citep{adejumo2026systematic, chen2025mitigating}. Moreover, most studies focus on isolated content generation rather than on structured workflows that can systematically enforce pedagogical coherence, institutional visual identity, and technical rigor across an entire curriculum. This gap motivates our proposal of an end-to-end architecture that transforms prompt engineering from an ephemeral, ad-hoc practice into a repeatable engineering process.

To address the reproducibility and precision gaps, we adopt the concept of Curriculum as Code, an approach that represents instructional materials as versionable, compilable, and reproducible artifacts through the use of programming and markup languages \citep{johnson2026toward, petullo2022courses, rodriguez2018courseware, clifton2007subverting}. In this framework, slides are not edited in WYSIWYG tools but written in LaTeX with Beamer, while technical figures are generated through Python scripts. This approach offers three fundamental advantages: first, mathematical expressions are rendered with typographic precision, mitigating errors common in standard presentation tools; second, the source code can be version-controlled, allowing for collaborative improvement and historical tracking; third, the compilation process automatically enforces a consistent visual identity, freeing the instructor from manual layout adjustments. Treating educational content as code creates an environment where automation becomes not only possible but also reliable.

A central challenge in designing such an architecture is the elicitation of the instructor's tacit pedagogical knowledge, i.e., the implicit rules, stylistic preferences, and contextual priorities that define his teaching signature \citep{wei2026grounding}. Knowledge elicitation, originally developed in the field of expert systems, involves externalizing such tacit expertise into explicit, formal rules. This tacit-explicit dichotomy, originally articulated by Polanyi \citep{polanyi1966tacit} and later extended to
organizational knowledge creation by Nonaka 
\citep{nonaka2008knowledge}, is a well-recognized challenge in knowledge engineering. The theoretical frameworks of expertise transfer emphasize that effective instructional systems must capture not only the declarative knowledge of the domain but also the procedural and conditional knowledge embedded in the instructor's choices regarding sequencing, emphasis, and contextualization \citep{swan2020toward}. The formalization of these pedagogical heuristics into explicit constraints is a prerequisite for any computational system aiming to replicate or assist the human instructional design process without losing pedagogical nuance.

Complementing this, Cognitive Load Theory (CLT) \citep{duran2022cognitive} provides the cognitive psychological foundation that dictates how information should be structured for optimal learning. CLT distinguishes between intrinsic cognitive load, determined by the complexity of the subject matter; extraneous cognitive load, caused by poor instructional design; and germane cognitive load, which facilitates schema acquisition. For engineering and STEM education, where the intrinsic complexity is inherently high, the primary theoretical implication is that instructional materials must drastically minimize extraneous load through clear structure, eliminate redundant information, and utilize signaling cues to guide attention. Consequently, a theoretically grounded instructional architecture must incorporate principles such as the modality effect, the split-attention effect, and the guidance-fading effect to ensure that the generated assets cultivate meaningful learning rather than cognitive overload. These theoretical directives establish the pedagogical boundaries within which any effective AI-assisted design workflow must operate.

\section{Materials and Methods}

\subsection{The Proposed Architecture: AI-Assisted Instructional Design Workflow}

The proposed architecture conceptualizes the creation of instructional materials as an engineering process. Drawing a direct analogy to Continuous Integration and Continuous Deployment (CI/CD) pipelines in software engineering, this workflow replaces unstructured human-AI interactions with a systematic, stage-gated pipeline. The architecture is composed of six sequential phases designed to enforce pedagogical coherence, technical precision, and visual consistency across the curriculum. 

\begin{figure}[t]
    \centering
    \includegraphics[width=0.9\linewidth]{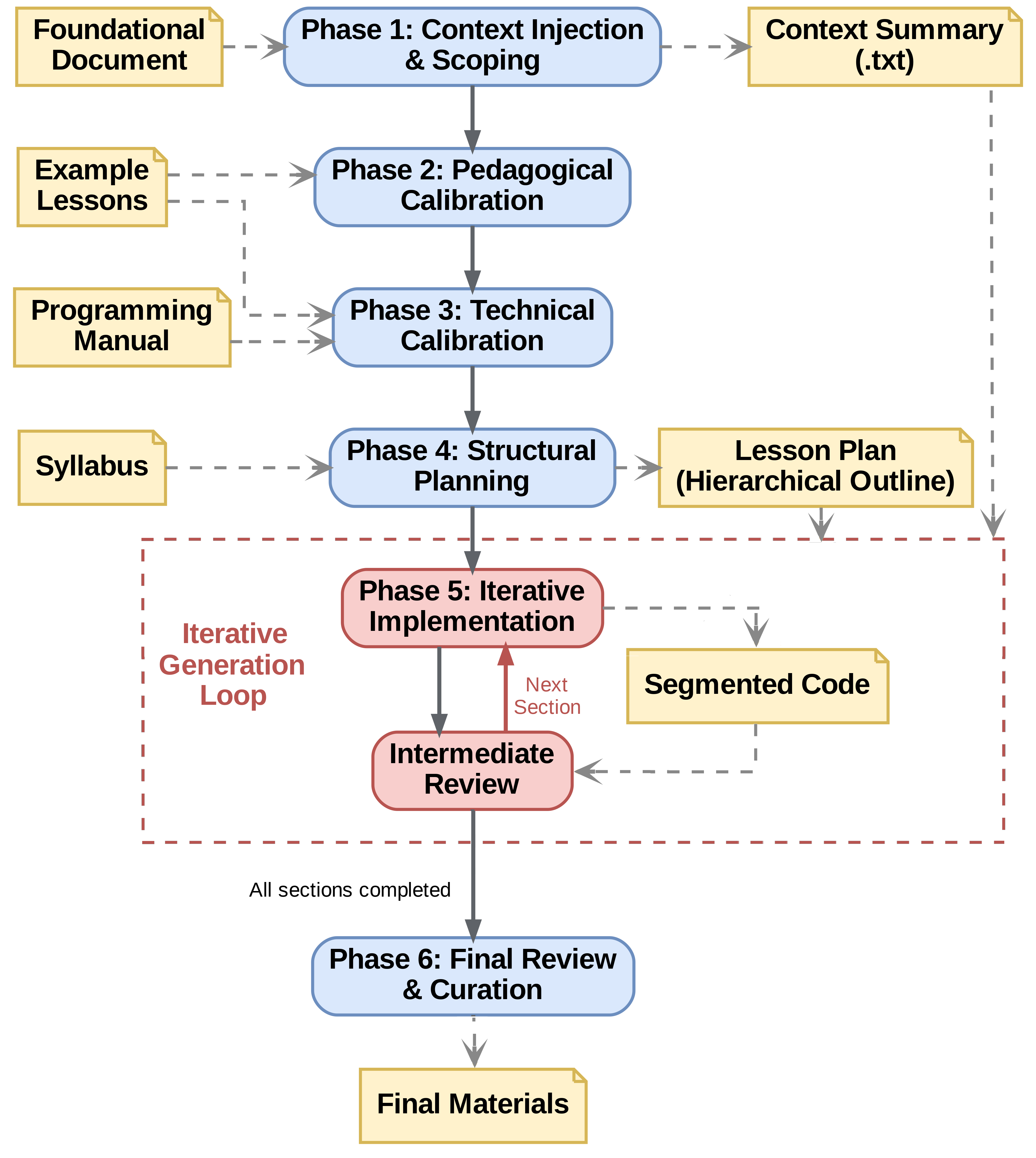}
    \caption{The six-phase AI-Assisted Instructional Design Workflow, illustrating prompt inputs, cyclical code generation (Phase 5), and human-in-the-loop validation nodes.}
    \label{fig:pipeline_diagram}
\end{figure}

To visually summarize this process, Figure \ref{fig:pipeline_diagram} illustrates the step-by-step diagram, detailing the prompt flows, the intermediate text-based artifacts, and the validation nodes at each stage.

\vspace{0.2cm}\noindent
\textbf{$\bullet$ Phase 1: Context Injection and Scoping.} The pipeline begins with the ingestion of a foundational document that dictates the lesson's scope. In traditional courses, this might be a syllabus or a textbook chapter summary. In highly contextualized environments, such as PBL, this is the Project Charter or a technical descriptive document. The LLM is tasked exclusively with understanding this input and generating a concise, three-paragraph textual summary (.txt) of the core concepts and constraints. An important technical design is that only the concise text artifact is carried forward to Phase 5. This deliberate forgetting mechanism prevents context window overflow and attention dilution, reducing the likelihood of hallucinations focus loss when generating complex code in later stages.

\vspace{0.2cm}\noindent
\textbf{$\bullet$ Phase 2: Pedagogical Calibration.} This phase focuses on eliciting and standardizing the instructor's tacit knowledge. Utilizing few-shot prompting with previously validated teaching materials, the AI is trained on the desired pedagogical style. This includes enforcing constraints such as cognitive load limits, time restrictions per slide, and the necessity to anchor theoretical concepts to practical applications. 

\vspace{0.2cm}\noindent
\textbf{$\bullet$ Phase 3: Technical Calibration.} Before any content is generated, the technical boundaries must be established. In this stage, the AI is injected with the specific syntax rules, required packages, and custom class calls required by the institution. This is intended to promote adherence of all subsequent outputs to the established visual identity and typographical standards, bypassing the need for manual formatting adjustments.

\vspace{0.2cm}\noindent
\textbf{$\bullet$ Phase 4: Structural Planning.} The AI generates a comprehensive lesson plan (a structural skeleton) aligned with the predefined learning objectives. To minimize cognitive load on the language model, no code is generated during this phase. The output is strictly a hierarchical outline defining sections, subsections, and the sequence of theoretical blocks and active learning exercises.

\vspace{0.2cm}\noindent
\textbf{$\bullet$ Phase 5: Iterative Implementation.} This phase operates as an algorithmic ``while'' loop. Attempting to generate the entire slide deck in a single prompt would inherently lead to token exhaustion and code truncation. To avoid these problems, the generation is heavily segmented. The AI is fed the structural skeleton from Phase 4 alongside the condensed context summary from Phase 1. It then iteratively generates the LaTeX code and the corresponding Python scripts (for data visualization or mathematical graphs) section by section. The loop continues until the entire structural plan has been implemented. By constantly referencing the Phase 1 summary, the model is much less likely to diverge into dense, generic theory, keeping the content strictly anchored to the specific lesson scope.

\vspace{0.2cm}\noindent
\textbf{$\bullet$ Phase 6: Review and Curation.} The final stage introduces a dual-layer validation process. Initially, the generated code and text undergo an automated review by a secondary, independent AI agent tasked with identifying syntax errors or pedagogical deviations. Following this automated curation, the primary instructor acts as the human-in-the-loop validator, compiling the code and approving the final material. In our empirical validation, to ensure rigor, this review was optionally extended to two independent peer instructors.

\subsection{Technical Implementation and Content Control}

To overcome the inherent limitations of standard WYSIWYG presentation tools and general-purpose AI outputs, this architecture operationalizes the \textit{Curriculum as Code} paradigm. The technical implementation relies on two primary computational tools: LaTeX (via the Beamer class) for document structure and typography, and Python for data visualization and mathematical graphing.

\vspace{0.2cm}\noindent
\textbf{a) Tooling and Visual Consistency:} The choice of LaTeX is justified by its deterministic nature and unparalleled precision in rendering complex mathematical equations and pseudocode. By injecting a custom institutional Beamer class during Phase 3, the architecture automatically enforces visual identity rules (e.g., color palettes, font sizes, and logo placement). This substantially relieves the instructor from manual layout adjustments. For graphical elements, the AI is prompted to generate Python scripts utilizing libraries such as Matplotlib and Seaborn. These scripts are strictly parameterized to output publication-quality figures 
, ensuring that all technical diagrams meet high academic standards.

\vspace{0.2cm}\noindent
\textbf{b) Pedagogical Constraint Enforcement:} Left unconstrained, Large Language Models tend to generate dense, verbose theoretical texts. To counteract this and adhere to Cognitive Load Theory principles, the architecture enforces strict pedagogical boundaries. The prompts in Phase 2 contain explicit rules dictating the maximum number of bullet points per slide, the fragmentation of long concepts, and the mandatory inclusion of applied activities. Most importantly, the AI is forced to continuously cross-reference the Phase 1 context summary, ensuring that every theoretical block is immediately followed by a practical scenario anchored in the project context (PBL). 

\vspace{0.2cm}\noindent
\textbf{c) Context Management and Interface Precision:} A critical design choice in this architecture is the restriction of the human-AI interface entirely to plain text, Markdown, and pure source code. This text-based protocol is highly precise and mathematically unambiguous. It helps avoid communication failures, formatting glitches, and unwanted escape characters that frequently occur when AI attempts to output rich-text or proprietary file formats. Furthermore, robust context management is achieved through the intentional token reduction established in Phase 1. By feeding the AI only the condensed text summary and the specific section plan during the Phase 5 loop, the model's attention mechanism remains sharply focused. This prevents the behavioral drift and code hallucinations that typically arise when an LLM's context window becomes overloaded with extensive project documentation.

\subsection{Experimental Setup and Empirical Validation}

To evaluate the robustness, scalability, and pedagogical fidelity of the proposed architecture, an empirical validation was conducted over a one-year academic period at the Institute of Technology and Leadership, a non-profit private higher education institution in Sao Paulo, Brazil \citep{paiva2025educational, lima2024teaching, brito2025pbl}. The institution operates strictly under a Project-Based Learning (PBL) paradigm structured in quarterly modules (trimesters). It offers a technology-oriented Business Administration (AD) program and four undergraduate computing programs: Computer Engineering (CE), Computer Science (CS), Information Systems (IS), and Software Engineering (SE).

The validation was divided into two distinct deployment scenarios to test both scalability across multiple instructors and adaptability across advanced specialized subjects:

\vspace{0.2cm}\noindent
\textbf{1. First-Year Common Core (High-Scalability Scenario):} The first year of the curriculum is common to all five undergraduate programs. The architecture was deployed across four consecutive quarterly modules, serving six distinct classes. Because each class undertakes a different practical project per module, the pipeline was subjected to a high-variability stress test, adapting the theoretical materials to 24 distinct project contexts throughout the year. The instructional materials for all these classes were generated by a single instructor (acting as the instructional designer) using the AI pipeline. To validate objective pedagogical fidelity, these materials were formally reviewed by two independent peer professors. Ultimately, the generated slide decks and exercises were successfully deployed in the classroom by six different professors, demonstrating the architecture’s capacity to produce high-quality instructional assets that remained effective across different instructors.

\vspace{0.2cm}\noindent
\textbf{2. Advanced Years (Specialization Scenario):} To verify the architecture's capability to handle highly specialized and dense technical constraints, it was also applied to five single-quarter modules in the second and third year. In these instances, the generated materials were adapted to one specific project per class and deployed solely by the authoring instructor.

\vspace{0.2cm} 
Throughout the year-long validation, the core generative tasks (Phases 1 through 5) were powered by institutional access to the Gemini Pro model family. To ensure rigorous validation during the automated curation stage (Phase 6), a secondary AI model, DeepSeek, was employed to cross-examine the generated code and text. Both platforms were utilized in their most recent versions available during the respective academic quarters (Gemini Pro 2.5 to 3.1 and DeepSeek 3).  

\section{Results}

This section reports the quantitative and qualitative outcomes of the one-year validation. The institution's curriculum is organized in quarterly modules numbered sequentially: modules 01–04 correspond to the first-year common core (shared by all five undergraduate programs), modules 05–08 to the second year, and modules 09–12 to the third year. The validation covered all four first-year common-core modules (IN01–IN04), each deployed across six distinct classes with different PBL projects, totaling 24 distinct project contexts. Additionally, four specialized modules were validated, representing second- (SE05, AD06) and third-year (CS11, SE12)  specializations. Tables~\ref{tab:core_tech} and~\ref{tab:core_proj} summarize, at a high level, the technical contents and the corresponding PBL projects for each module involved in this study. 

\begin{table}[t]
\begin{center}
\caption{Technical contents of each module.}
\label{tab:core_tech}
\begin{tabular}{p{8.2cm}}
\hline
\textbf{IN01:} Functions \& limits; derivatives (basic rules); derivative applications (optimization, L'Hôpital); coordinate systems \& vectors; kinematics (1D/2D); dynamics, energy \& momentum;  \\
\textbf{IN02:} Indefinite integrals; definite integrals \& FTC; descriptive statistics; propositional logic;  numerical methods (bisection, Newton, integration); graphs, trees \& coloring. \\
\textbf{IN03:} Multivariable functions; partial derivatives; multiple integrals; Probability distributions; inductive statistics (CLT, CI, hypothesis testing); linear transformations. \\
\textbf{IN04:} Vector fields (grad, div, curl); electric field \& potential; directional derivatives; power/Taylor series; circuits (resistors/capacitors, Kirchhoff laws); EM waves. \\
\hline
\textbf{SE05:} Single \& multivariable differentiation; optimization (Hessian, extrema); L'Hôpital; applications to software systems (cost, latency, load balancing). \\
\textbf{AD06:} Descriptive stats; Bayes theorem; probability distributions (binomial, uniform, exponential, normal, log-normal, triangular); Monte Carlo simulation. \\
\hline
\textbf{CS11:} Linear algebra (BoW, embeddings, similarity); entropy \& information gain; time series (stationarity, smoothing, ARIMA). \\
\textbf{SE12:} Propositional logic; linear programming (formulation, graphical solution, simplex, duality). \\
\hline
\end{tabular}
\\[0.1cm]
\end{center}
\textbf{IN:} Initial (first-year common track), \textbf{AD:} Business Administration, \\ 
\textbf{CS:} Computer Science, \textbf{SE:} Software Engineering
\vspace{-0.2cm}
\end{table}

\begin{table}[t]
\begin{center}
\caption{PBL project contexts of each module}
\label{tab:core_proj}
\begin{tabular}{p{8.2cm}}
\hline
\textbf{IN01:} Game development: financial education, pet store engagement, museum gamification, eye-tracker accessibility, sales training, and logistics. \\
\textbf{IN02:} Platforms: remote assessment, civil defense georeferencing, livestock management, sports event tracking, agricultural governance, educational monitoring. \\
\textbf{IN03:} Predictive models: healthcare demand, animal protein, vehicle sales, agricultural diesel, NPS detractors, and thermal forecasting. \\
\textbf{IN04:} Monitoring systems: in-car noise, domestic boiler, electrical transformer, and access control for high-speed trains. \\
\hline
\textbf{SE05:} SaaS onboarding automation (data-driven configuration, cost/latency optimization, load balancing). \\
\textbf{AD06:} Agricultural data modeling and decision-making (crop optimization, risk management, value at risk). \\
\hline
\textbf{CS11:} Intelligent legal assistant (semantic understanding, classification, operational monitoring, forecasting). \\
\textbf{SE12:} Generic real-world business/industry/logistics problems (this particular module had no fixed PBL project). \\
\hline
\end{tabular}
\\[0.1cm]
\end{center}
\textbf{IN:} Initial (first-year common track), \textbf{AD:} Business Administration, \\ \textbf{CS:} Computer Science, \textbf{SE:} Software Engineering  
\vspace{-0.2cm}
\end{table}

\begin{table}[t]
\begin{center}
\caption{Student assessment about the material}
\label{tab:student_assessment}
\begin{tabular}{ccc}
\hline  
Module & Assessment (mean $\pm$ std) & $n$ \\
\hline  
IN01 & $8.5 \pm 0.5$ & $219$ \\
IN02 & $9.1 \pm 0.5$ & $207$ \\
IN03 & {\textit{Pending}} & -- \\
IN04 & $9.5 \pm 0.3$ & $118$ \\
\hline
SE05 & $9.8 \pm 0.8$ & $29$ \\
AD06 & $9.8 \pm 0.5$ & $18$ \\
\hline
CS11 & {\textit{Pending}} & -- \\
SE12 & $9.9 \pm 0.6$ & $13$ \\
\hline
\end{tabular}
\end{center}
\textbf{IN:} Initial (first-year common track), \textbf{AD:} Business Administration, \\ \textbf{CS:} Computer Science, \textbf{SE:} Software Engineering  
\vspace{-0.2cm}
\end{table}

\subsection{Implementation Scale and Role Distribution}
A critical aspect of evaluating the architecture's scalability was separating the instructional design from the classroom delivery. For the first-year common core (IN01--IN04), all instructional materials were generated by a single instructor who acted as the instructional designer, operating the AI pipeline. To ensure objective quality, these materials were systematically peer-reviewed by two independent professors before deployment. Ultimately, the generated slide decks and Python-based activities were utilized in the classroom by six different professors. This separation provided evidence that the generated assets were applicable across different instructors and were not overly dependent on the creator’s personal delivery style, suggesting that tacit knowledge had been successfully translated into reusable teaching assets.

\subsection{Time Efficiency and the Human-in-the-Loop}
The deployment of the architecture was associated with a strictly quantifiable reduction in material preparation effort. Methodologically, the time required to author a complete modular technical slide deck was tracked and compared against historical baselines from previous academic years.

Traditionally, developing customized PBL slide decks—complete with complex mathematical typesetting, accurate Python plots, and institutional formatting—required an average of eight hours of continuous manual effort per instruction. With the proposed pipeline, the heavy lifting of coding LaTeX and Python was entirely offloaded to the AI. Consequently, the active human-in-the-loop workload dropped to an average of two hours per instruction.

This remaining human time is strategically distributed: approximately 30 minutes are spent in Phase 4 (Structural Planning), where the instructor reviews and tweaks the text-based Lesson Plan. The remaining 90 minutes are dedicated to overseeing the iterative generation loop (Phase 5) and performing the Final Review \& Curation (Phase 6), focusing on pedagogical adjustments and didactic pacing rather than syntax troubleshooting. Because the AI strictly followed the approved plan during the subsequent coding phase, the most time-consuming tasks were automated, yielding an estimated 75\% reduction in preparation time—without sacrificing the instructor’s personal teaching signature.

\subsection{Technical Fidelity and Code Hallucinations}
Technical fidelity was evaluated based on the AI's ability to produce compilable LaTeX code and executable Python scripts on the first attempt (zero-shot within the generation loop). The architecture demonstrated high reliability during the observed deployments. Constrained by the strict parameterization enforced during Phase 3, the Python scripts generated for data visualization invariably executed without compilation errors on the zero-shot attempt, successfully outputting publication-quality figures (e.g., adhering to \texttt{dpi=300} and specific library constraints).

Regarding LaTeX generation, the system achieved extremely high rates of both semantic content fidelity and syntactic compilation success. The intentional context pruning in Phase 1 and the isolated section-by-section generation loop in Phase 5 effectively eliminated token exhaustion. As a direct result of this architectural constraint, no conceptual or mathematical hallucinations were identified during the human-in-the-loop review across the 28 project contexts. The very few compilation errors encountered were minor syntactic anomalies, the most common being mismatched bracket closures (e.g., outputting \texttt{\{]} instead of \texttt{\{\}} or \texttt{[]}). These superficial errors were easily caught by the automated review agent (Phase 6) and instantly corrected, proving that the multi-phase, segmented architecture is highly resilient against the typical failure modes of Large Language Models.

The robustness of the architectural constraints was further corroborated by the independent peer-review process prior to classroom deployment. Peer reviewers consistently reported a complete absence of conceptual or mathematical hallucinations in the generated materials, confirming the accuracy of complex calculations and derivations. Instead of correcting technical AI failures, the human-in-the-loop interventions were strictly elevated to the level of pedagogical curation. Review revisions were limited to minor didactic enhancements, such as explicitly expanding intermediate calculation steps in answer keys for better student comprehension, refining the theoretical boundaries of physical constants (e.g., distinguishing between maximum and variable static friction). This shift confirms that the segmented generation loop successfully isolates the LLM from context overload, yielding a highly reliable baseline that only requires final stylistic and pedagogical polishing.

\subsection{Pedagogical Fidelity and Visual Consistency}
The architecture consistently reflected the pedagogical constraints required by the PBL methodology. Across all generated modules, the AI strictly adhered to the mandated proportion between theoretical slides and practical applications. Theoretical concepts were systematically fragmented to avoid cognitive overload and were consistently followed by project-anchored activities and mathematically dense solution slides. 

\begin{figure}[t]
    \centering
    (a)\vspace{0.1cm}
    \includegraphicsborder[width=1.0\linewidth]{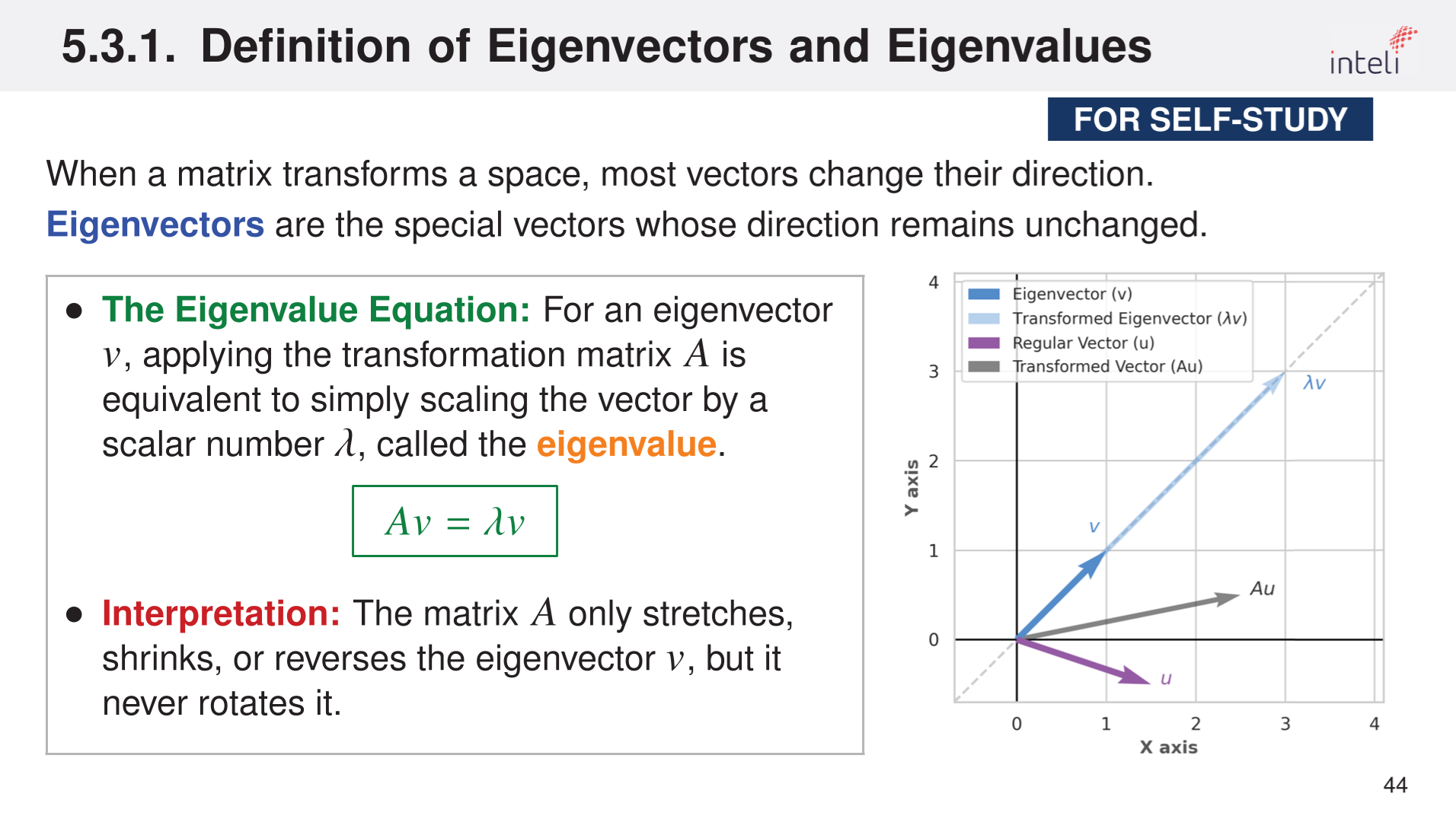} 
    (b)\vspace{0.1cm}
    \includegraphicsborder[width=1.0\linewidth]{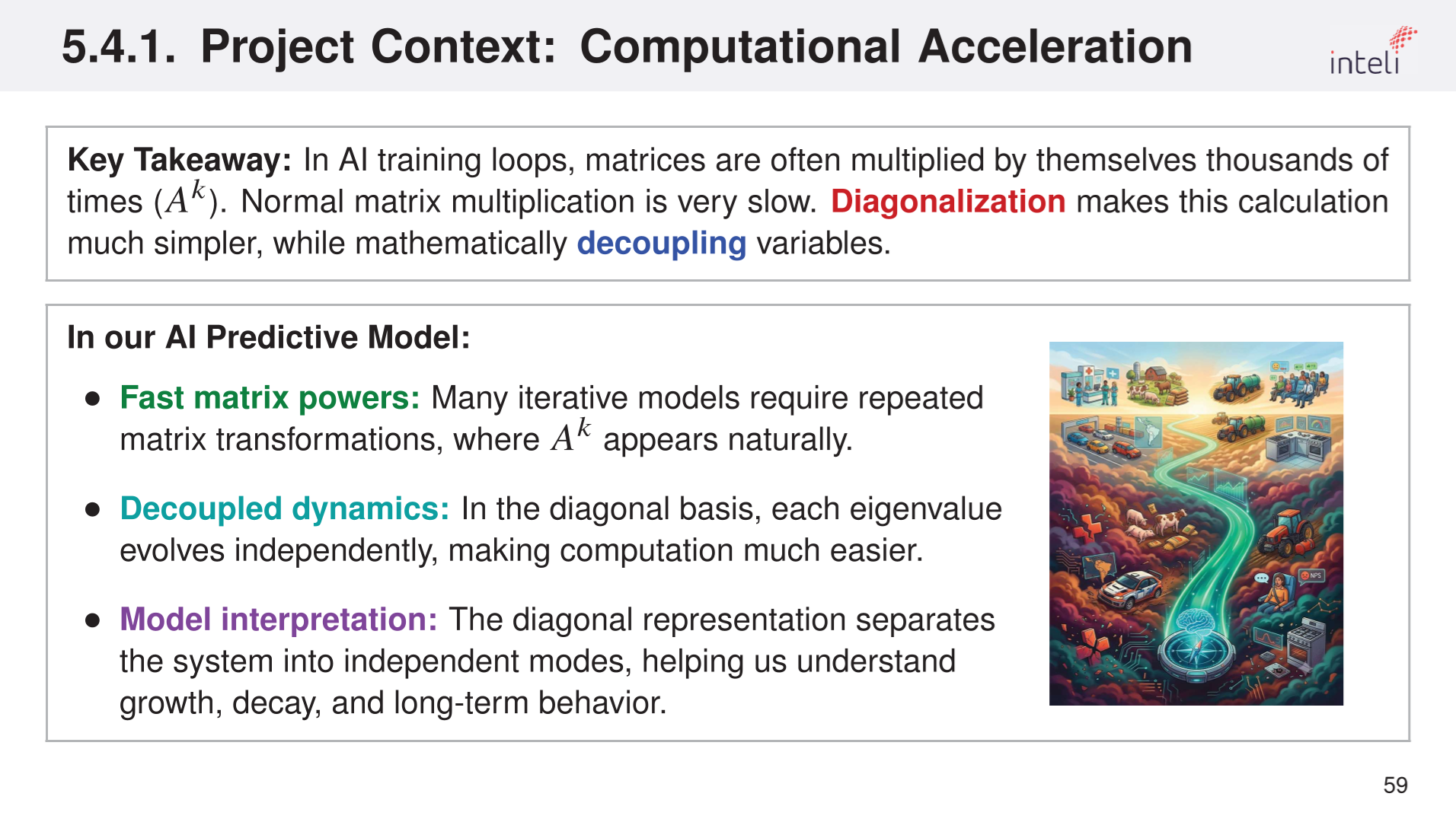} 
    (c)\vspace{0.1cm}
    \includegraphicsborder[width=1.0\linewidth]{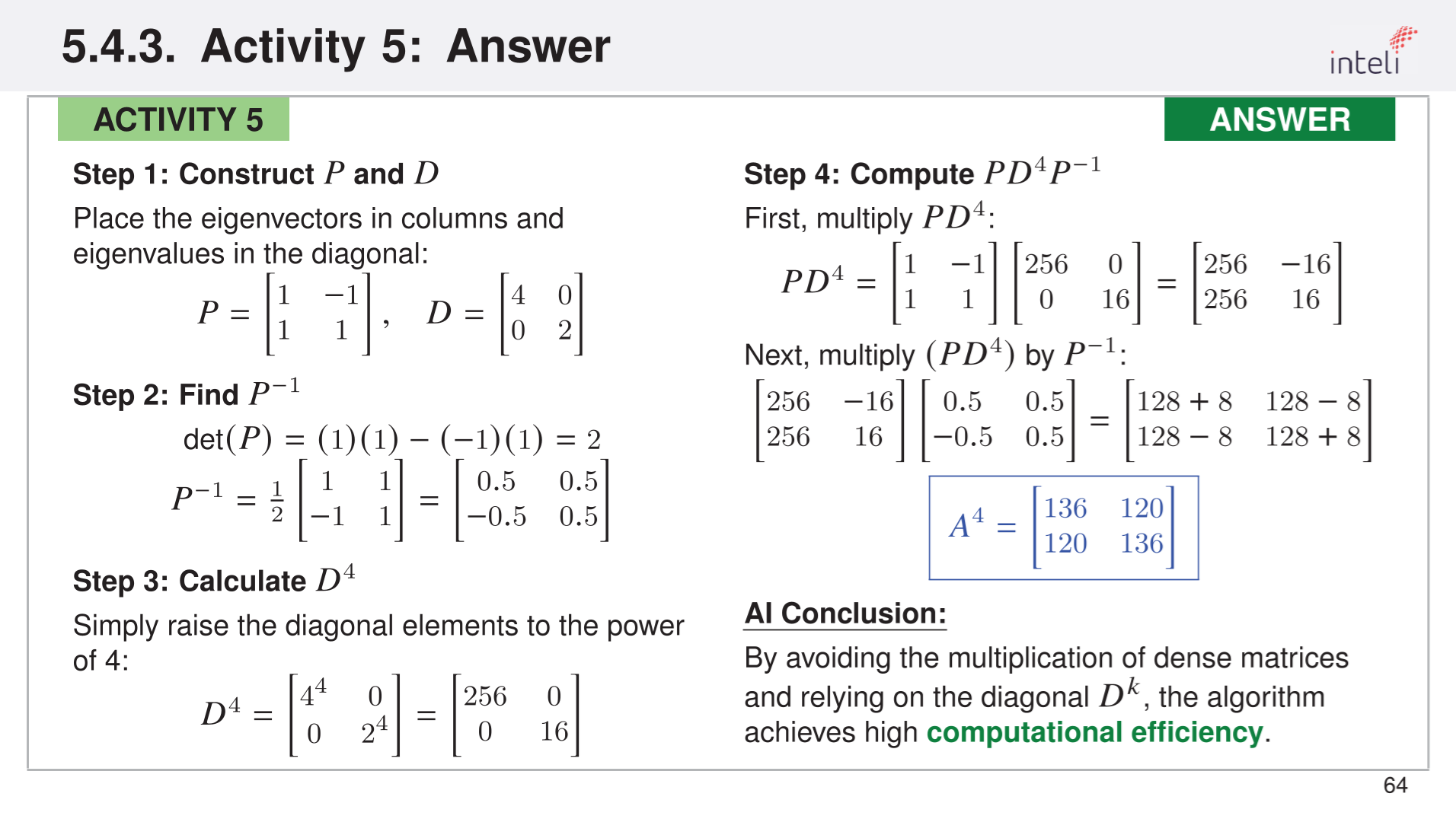} 
    \caption{Examples of generated materials demonstrating visual consistency and technical precision: (a) Slide with Python-generated data visualization, (b) Project-anchored application, and (c) Mathematically dense solution slide.}
    \label{fig:slide_examples}
\end{figure}

Figure \ref{fig:slide_examples} illustrates three representative slides generated entirely by the pipeline: a contextualized theory application, a practical PBL activity, and a mathematically dense answer slide. Although the primary language of instruction during the validation period was Portuguese, the architecture was successfully employed to generate identical English versions to accommodate exchange students, as depicted in these translated, publication-ready examples. Furthermore, because the formatting was handled entirely by a custom institutional Beamer class injected in Phase 3, the system seamlessly managed bilingual content without breaking the layout. Consequently, the visual identity remained fully cohesive across all languages and disciplines, independent of which of the six professors was delivering the class.

\subsection{Qualitative Impact: Student Assessment and Cognitive Load}
To evaluate the end-user impact, anonymous voluntary student feedback was collected at the end of each quarterly module. In a regular survey on the  sessions, students were asked to rate six statements about the lecturers, concerning six criteria: planning, material, didactics, rapport, availability and feedback. The specific statement about material is \textit{"The provided material is of high quality and understandable"}, rated on a scale from 0 to 10. 

Table \ref{tab:student_assessment} presents the results (mean $\pm$ standard deviation) regarding the quality of the material, as well as the number of assessments per module (which were anonymous and voluntary, therefore lower than the total number of students in the class). There is significantly more participation on the first year (IN01-IN04) because of the number of classes involved. 

It is worth noting that while the instructional materials for modules IN03 and CS11 have been successfully generated and have passed the human-in-the-loop validation (Phase 6) without technical hallucinations, their classroom deployment is currently taking place. Consequently, the final student evaluation metrics for these specific modules are pending and will be consolidated upon the completion of the academic term.

As detailed in Table \ref{tab:student_assessment}, the generated materials received consistently high ratings, ranging from an average of $8.5$ to $9.9$ across both foundational and advanced modules. These results are consistent with Cognitive Load Theory (CLT). By systematically enforcing content fragmentation, eliminating redundant text, and maintaining a predictable visual organization and high typographic quality through LaTeX, the architecture was designed to reduce students’ extraneous cognitive load. The consistently high ratings indicate that students perceived the materials as high-quality and understandable. This interpretation is further supported by the independent review and approval of the generated materials by the course instructors.

\section{Discussion}

The results of this year-long implementation provide substantial evidence for the viability of automated, code-driven instructional design. By structuring the interaction between the instructor and the Generative AI, the proposed architecture effectively addressed the core challenges of technical reproducibility and pedagogical coherence.

\subsection{Synthesis of Findings}
Empirical validation provided evidence addressing the three research questions that guided this study.

\vspace{0.2cm}\noindent
\textbf{$\bullet$ RQ1 (Pedagogical Framework):} The architecture suggests that tacit pedagogical knowledge can be systematically translated into explicit rules through a combination of few-shot calibration (Phase 2) and structural isolation (Phase 4). By allowing the instructor to quickly review and tweak the text-based Lesson Plan before any code was generated, the pipeline ensured that the instructor's personal teaching signature was preserved while automating the most labor-intensive tasks.

\vspace{0.2cm}\noindent
\textbf{$\bullet$ RQ2 (Technical Reproducibility):} The integration of code-based tools (LaTeX/Beamer and Python) with Generative AI mitigated the problem of mathematical hallucinations. The critical mechanism enabling this was the drastic context pruning in Phase 1 and the iterative generation loop in Phase 5. Feeding the model only isolated sections of the lesson plan alongside a strictly condensed project summary kept the AI's attention focused, ensuring highly accurate code outputs.

\vspace{0.2cm}\noindent
\textbf{$\bullet$ RQ3 (Practical Impact):} The practical deployment confirmed a significant reduction in faculty preparation time. Furthermore, because the visual formatting was centralized within a custom Beamer class, the institutional visual identity was maintained across 24 distinct projects and multiple disciplines. The successful delivery of these materials by six different professors provided evidence that the assets could be adopted across multiple instructors, supporting their scalability.

\subsection{Beyond Prompt Engineering: The Primacy of Architecture}
A critical observation from the validation phase is that the exact phrasing of the prompts or the specific foundational model used (e.g., Gemini or DeepSeek) had a secondary impact on the final quality compared to the structure of the workflow itself. The current literature heavily emphasizes prompt engineering, but our findings suggest a paradigm shift towards instructional architecture. 

The strength of this architecture lies in its constraints: the drastic pruning of context, the isolation of structural planning from code generation, and the iterative loop that implements one section at a time. Attempting to generate a full slide deck in a single prompt inevitably leads to token exhaustion and hallucinations. Furthermore, restricting the human-AI interface to pure text, markdown, and source code proved to be a highly precise communication protocol, successfully avoiding the formatting glitches common when AI models attempt to output rich text or proprietary file formats.

\subsection{Practical Implications}
This framework promotes a fundamental evolution in the role of the educator. The instructor transitions from being a manual formatter of slides to acting as an architect and engineer of instructional design. 

The broad applicability of the proposed solution should be emphasized. The architecture was deliberately stress-tested in a Project-Based Learning (PBL) environment, which adds a severe layer of complexity due to the continuous need to anchor theoretical concepts to varying practical projects. Because the pipeline successfully resolved this highly demanding scenario, its application to traditional, lecture-based courses across diverse higher education institutions becomes straightforward and highly scalable. The pedagogical style enforced is not a rigid requirement; the architecture is entirely adaptable to any instructional methodology.

A  highly valuable implication of this architecture is its native support for internationalization. Because the content is generated strictly as code (text-based), utilizing the LLM to translate entire modules for exchange students proved trivial. Unlike traditional WYSIWYG tools, where changing the language often disrupts the visual layout, the compiled LaTeX code automatically accommodates text expansion or contraction, maintaining perfect visual consistency across different languages.

\subsection{Challenges and Limitations}
Despite its successes, the implementation faced practical challenges. Generative models are subject to behavioral drift following systemic updates by their providers, occasionally requiring minor recalibrations in the prompt instructions. Additionally, while conceptual hallucinations were eliminated, minor syntactic anomalies still occurred, such as the occasional mismatch of brackets in LaTeX. 

While the first-year foundational modules successfully demonstrated the architecture's high scalability across multiple instructors and large student cohorts ($n > 200$), the advanced specialization modules (Years 2 and 3) were limited to smaller class sizes and single-instructor deployment. On the other hand, these advanced modules were required for stress-testing the architecture's capacity to handle highly dense, specialized mathematical content without hallucinations.

The primary limitation of this study is that, although the generated materials were deployed across multiple courses and utilized by various professors, the authoring process was conducted by a single instructor within a single institution. Future multisite studies are necessary to observe how different faculty members adapt to operating the pipeline.

\subsection{Future Work}
The modular nature of the proposed architecture (as illustrated in Figure 1) was deliberately designed to directly support the integration of autonomous AI Agents and Skills. Because the pipeline is already strictly segmented into discrete phases with clear inputs and outputs, future research should focus on replacing the manual web-based AI interface with fully automated orchestration via APIs, deploying specialized agents for each specific phase. Integrating this agentic pipeline with Git repositories would enable a CI/CD-inspired workflow for educational content development, supporting automated generation, versioning, validation, and deployment.

Finally, the scalability of this architecture opens avenues for advanced customization. Future iterations of the pipeline could automatically generate variations of the same material adapted to different student learning profiles (adaptive learning) and produce accessible versions for visually impaired students.

\section{Conclusion}

This paper presented a robust and scalable AI-assisted architecture for instructional design, specifically tailored to the rigorous demands of STEM and engineering education. By formalizing the creation of educational materials into a structured, multi-phase pipeline grounded in the Curriculum as Code paradigm, this approach successfully transcends the ephemeral and error-prone nature of ad-hoc prompt engineering. The empirical validation demonstrated that the deliberate segmentation of tasks, combined with strict context pruning and the use of precise code-based tools like LaTeX and Python, effectively mitigates the well-documented limitations of Large Language Models, such as mathematical hallucinations and formatting inconsistencies.

One of the main contributions of this work is the systematization of the instructor's tacit pedagogical knowledge. The proposed architecture translates individual teaching practices and complex requirements into explicit, replicable computational rules, allowing institutions to reduce faculty workload while preserving instructional quality. As active learning methodologies increasingly require customized and contextualized educational resources, engineering-oriented frameworks of this kind can provide a structured approach to content development. Ultimately, the proposed approach supports the scalable production of high-quality educational materials, allowing instructors to devote more time to activities that benefit most from their expertise, such as facilitating student learning and innovation.

\section*{Acknowledgments}

The author acknowledges the institutional support of Unifesp and Inteli. The independent review of the AI-generated educational material by Geraldo Magela Severino Vasconcelos and Diogo Morais is also gratefully acknowledged.

\section*{Statements}
The author has no competing interests to declare that they are relevant to the content of this article.

During the preparation of this work, the author used Gemini 3.1 Pro for drafting support and language polishing of the manuscript text. All content was thoroughly reviewed, verified, and edited by the author, who takes full responsibility for the final text.

{
\fontsize{9}{10}\selectfont

}

\end{document}